\def\year{2020}\relax
\documentclass[letterpaper]{article} 
\usepackage{aaai20}  
\usepackage{times}  
\usepackage{helvet} 
\usepackage{courier}  
\usepackage[hyphens]{url}  
\usepackage{graphicx} 
\usepackage{amsmath}
\usepackage{graphicx}  
\title{DeepVar: An End-to-End Deep Learning Approach for \\ Genomic Variant Recognition in Biomedical Literature}

\author{Chaoran Cheng,\textsuperscript{\rm 1}
Fei Tan,\textsuperscript{\rm 2}
Zhi Wei\textsuperscript{\rm 1}\thanks{Corresponding Author}\\ 
\textsuperscript{\rm 1}New Jersey Institute of Technology, Newark, NJ, USA \\ 
\textsuperscript{\rm 2}Yahoo Research, New York, NY, USA \\
cc424@njit.edu, fei.tan@verizonmedia.com, zhi.wei@njit.edu
} 
\begin{document}

\maketitle

\begin{abstract}
We consider the problem of Named Entity Recognition (NER) on biomedical scientific literature, and more specifically the genomic variants recognition in this work. Significant success has been achieved for NER on canonical tasks in recent years where large data sets are generally available. However, it remains a challenging problem on many domain-specific areas, especially the domains where only small gold annotations can be obtained. 
In addition, genomic variant entities exhibit diverse linguistic heterogeneity, differing much from those that have been characterized in existing canonical NER tasks. The state-of-the-art machine learning approaches heavily rely on arduous feature engineering to characterize those unique patterns. 
In this work, we present the first successful end-to-end deep learning approach to bridge the gap between generic NER algorithms and low-resource applications through genomic variants recognition. 
Our proposed model can result in promising performance without any hand-crafted features or post-processing rules. Our extensive experiments and results may shed light on other similar low-resource NER applications. 
\end{abstract}

\section{Introduction}
Due to the up-surging volume of new biomedical literature in the past decades, it becomes infeasible for researchers to access all those up-to-date publications manually. The automated information extraction tools play a critical role in assisting researchers to keep up with the explosive knowledge effectively. In general, the first step is to identify name entities from text, which is termed as Named Entity Recognition (NER), a common task in the nature language processing field. 
In the biomedical context, entities are typically short phrases as the representations of a specific object, e.g., names of genes or proteins, genetic variants, diseases, drugs, etc. Moreover, a noticeable amount of those entities contain letters, digits, and punctuation, resulting in more complex semantic alternations and differing much from entities characterized in news or conventional articles.

To identify named entities present in the text, statistical approaches such as Maximum Entropy (ME) 
and Conditional Random Fields (CRFs) 
are used in most of the previous works with either learning patterns associated with a particular type of entities or hand-built rules. The performance of such algorithms heavily depends on the design of hand-crafted features. Recently, the Deep Neural Network (DNN) models have increasingly been used in generic NER tasks and achieved significant success, pushing most of the benchmarks to a new level. 
More importantly, those models minimized the feature engineering efforts by learning the hidden patterns from a large volume of labeled samples.

\begin{table*}[t]
\caption{Comparison of Other Data Sets with Ours to Demonstrate The Extreme Low-Resource Situation in Our Work} \smallskip
\label{data_comp}
\centering
\resizebox{0.95\textwidth}{!}{ 
\begin{tabular}{l|r|r|r}
\hline
\multicolumn{1}{c|}{\textbf{Data Set}} & \multicolumn{1}{c|}{\textbf{Size}} & \multicolumn{1}{c|}{\textbf{Entity types and counts}} &
\multicolumn{1}{c}{\textbf{Named Entity Example}}\\ \hline
BC4CHEMD & 47,402 sentences & Chemical (84,310) & (25)MgPMC16; SAHA\\ \hline
BC5CDR & 30,677 sentences & \begin{tabular}[c]{@{}r@{}}Chemical(15,935)\\ Disease(12,852)\end{tabular} & 
\begin{tabular}[c]{@{}r@{}}cyclosporin A; L-dopa\\ cardiovascular arrhythmias; swelling\end{tabular}\\ \hline
BC2GM & 20,000 sentences & Gene/Protein (24,583) & S-100; Cdc42; RecA; ROCK-I\\ \hline
JNLPBA & 13,484 sentences & \begin{tabular}[c]{@{}r@{}}Cell Line(4,330) \\ DNA (10,589)\\ Gene/Protein (20,448) \\ Cell Type (8,649) \\ RNA(1,069)\end{tabular} & 
\begin{tabular}[c]{@{}r@{}}Jurkat T-cells; Hsp60-specific T cells\\ cytokine gene; human interleukin-2 gene;\\ NF-kappaB site; Hsp60; retinoic acid receptors \\16HBE human bronchial epithelial cells \\ GR mRNA; glucocorticoid receptor mRNA\end{tabular}\\ \hline
NCBI-Disease & 8,336 sentences & Disease(6,881) &  MCF-7 tumours; breast and ovarian cancer \\ \hline
\textbf{tmVar - Ours} & 4,783 sentences & \begin{tabular}[c]{@{}r@{}}Protein Mutation (653)\\ DNA Mutation (751) \\ SNP (136)\end{tabular} & 
\begin{tabular}[c]{@{}r@{}}p.Pro246HisfsX13; S276T; Arg987Ter\\ c.399\_402del AGAG; Ex2+860G$>$C; -866 promoter(G/A); \\ rs2234671; rs1639679\end{tabular}\\ \hline
\end{tabular}
}
\end{table*}

Our goal in this work is to develop an end-to-end DNN NER model that can automatically identify variants in biomedical literature and classify them into a set of predefined types. However, due to the prohibitive cost of expert curation, the size of curated training data with gold label annotations is often restricted in biomedical domains. As shown in Table \ref{data_comp}, the sample size of the benchmark dataset of variants, tmVar \cite{wei2013tmvar}, is much smaller than others. Furthermore, it exhibits more exotic linguistic heterogeneity. 
The complex morphological heterogeneity exacerbates the challenge for solving this problem, let alone the small data size. Despite numerous attempts on other biomedical benchmarks in the past, it is the first attempt to leverage a deep learning approach for the genomic variants recognition. The main challenges in this work include:
\begin{itemize}
    \item To minimize feature engineering effort, automatically generalizing hidden diverse linguistic patterns is harder from limited training resources.
    \item To differentiate the ambiguous entities or synonym, learning some effective feature representation is harder with shallow networks from restricted resources.
    \item To limit the false positive error, both the entity identification and the entity boundaries need to be accurately inferred, which is critical for downstream applications such as entity normalization and relation extraction.
\end{itemize}

In this work, we took full advantage of the generic state-of-the-art deep learning algorithms and proposed a Deep Variant (DeepVar) Named Entities Recognition model. We aimed to find a principled way to transfer domain knowledge and build an end-to-end DeepVar model. Our results show that DeepVar could achieve better performance than state-of-the-art algorithms using significantly less domain knowledge and without any feature engineering.

\begin{figure*}[!t]
    \centering
    \includegraphics[width=0.7\textwidth, height=0.5\textwidth]{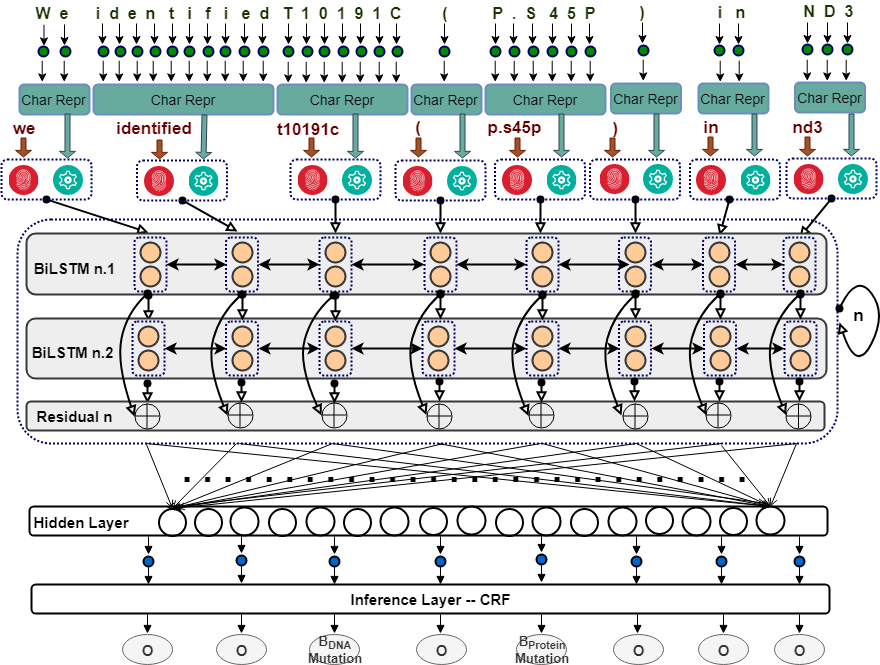}
    \caption{The Architecture of Proposed DeepVar NER Model. The green color module on top represent character input and sequence representation learning; the red circle icon represents word embedding; the gray boxes represent word sequence representation learning module including multiply BiLSTM and Residual layers. The sentence "We identified T10191C(P.S45P) in ND3." used in the figure is for illustration purpose. }
    \label{fig:model}
\end{figure*}

\section{Related Works}\label{sec:related}
In this section, we summarized the recent works in generic NER and the recent efforts of applying generic NER algorithms to the biomedical domain. We also outlined the recent works in genomic variants recognition.

\noindent\textbf{Generic NER.} In earlier works, statistical machine learning systems have proven its success for NER \cite{nadeau2007survey,nothman2009analysing} with various feature engineering efforts like building internal linguistic features. More recently, due to the development of deep learning techniques, it becomes a fashion in NER applications to minimize the efforts of feature engineering and build an end-to-end system. The first attempt to use deep learning in NER task should be the SENNA system \cite{collobert2011natural}, which still utilized lots of hand-crafted features. 
The current state-of-the-art approaches regulate both the word level and character level representations intertwined by both bidirectional Long Short-Term Memory (LSTM) \cite{hochreiter1997long} Neural Network 
and Convolutional Neural Network (CNN) \cite{zeiler2010deconvolutional}. 
Some works focused on building the shallow word-level representations with character-based feature through CNN \cite{collobert2011natural,zhang2015character,kim2016character,strubell2017fast}, or bidirectional LSTM (BiLSTM) \cite{ma2016end,strubell2017fast}. 
The majority of works combined both word-level and character-level features to achieve the best performance. Nevertheless, some still applied slight pre-processing steps like normalizing digit characters, while some works employed marginal hand-crafted features to some extent \cite{chiu2016named,strubell2017fast}. 
\citeauthor{lample2016neural} \shortcite{lample2016neural} and \citeauthor{ma2016end} \shortcite{ma2016end} achieved the end-to-end manner without any hand-crafted effort. 

\noindent\textbf{BioNER.} The Biomedical Named Entity Recognition (BioNER) tasks focus on extracting biomedical domain entities such as cell lines, diseases, genes, and proteins. 
Various similar machine learning-based approaches have also been applied in earlier works and achieved good performance. The widely used hand-crafted features include different types of linguistic features such as orthographic features, word-shape features, n-gram features, dictionary features, and context features, as well as domain-specific features from biomedical terminologies. Various works have been done on several BioNER tasks to prove the effectiveness of aforementioned models. 
\citeauthor{habibi2017deep} \shortcite{habibi2017deep} investigated the effectiveness of approach proposed in \cite{lample2016neural} for chemicals, diseases, cell lines, species, and genes name recognition, while \citeauthor{dernoncourt2017identification} \shortcite{dernoncourt2017identification} verified the same approach on patient notes. \citeauthor{yoon2019collabonet} \shortcite{yoon2019collabonet} investigated the approaches of \cite{ma2016end} on chemicals and disease entities. \citeauthor{wang2018cross} \shortcite{wang2018cross} utilized the similar multitask architecture in \cite{liu2018empower} and verified on chemicals, cell lines, disease, genes, and other name recognition. \citeauthor{xu2018improving} \shortcite{xu2018improving} proposed a modified framework based on \cite{lample2016neural} by adding extra sentence level representation as global attention information and verified on clinical NER task. 
Nevertheless, some recent works still need to elaborate on marginal external information.

\noindent\textbf{Genomic Variants Recognition.} With respect to the genomic variants recognition, all the previous works including MutationFinder \cite{caporaso2007mutationfinder}, EBNF \cite{laros2011formalized}, OpenMutationMiner \cite{naderi2012automated}, tmVar \cite{wei2013tmvar}, SETH \cite{SETH2016}, and NALA \cite{cejuela2017nala} employed dozens of regular expressions to build orthographic and morphological features, like word shape, prefixes, and suffixes, for their variants entities identification systems. 
Since the regular expressions used for generating customized hand-crafted features are fixed and can only describe limited patterns, all the previous works mainly focused on techniques improving the regular expressions to capture more patterns \cite{naderi2012automated,wei2013tmvar,SETH2016}. Nevertheless, they still need to add a bunch of post-processing steps to achieve better results \cite{wei2013tmvar,cejuela2017nala}. Moreover, despite the efforts in recent BioNER tasks, the leverage of deep learning approach in variant identification tasks remains open in literature, and to build an end-to-end approach can be challenging.

\noindent\textbf{Word Embedding.} It's worth noting that most of those works employed the word-level distributional representations, well-known as word embedding. 
However, the pre-trained embeddings on generic corpora cannot fully capture the semantics of biomedical entities. One of the common challenges is the OOV words, which can be rare terms like mutants or unseen forms of known words like chemical names. Those entities are not typo and have high occurrence but cannot be found in the canonical pre-trained embeddings. Recently, the word representations pretrained on a large collection of domain-specific texts (PubMed, PMC, etc.) are proved its superiority than generic word embeddings \cite{habibi2017deep,mohan2018fast}. 
Moreover, the most recent proposed contextual embeddings such as ELMO \cite{Peters2018ELMo}, Flair \cite{akbik2019flair}, and BERT \cite{devlin2019bert} achieved state-of-the-art performance on all the generic NLP applications. Their domain-specific embeddings (BioELMO, BioFlair, and BioBert), which are trained on the large biomedical corpus, are simultaneously made available to the public. Various works \cite{jin2019probing,peng2019transfer} showed that they outperform word2vec \cite{mikolov2013distributed} on BioNER tasks. In our experiments, we also investigated different pre-trained BioEmbeddings. 

\section{Deep Variants Identification Model}

In this section, we presented our DeepVar NER model for identifying variants in a low-resource data set. We focused on neural sequence representation learning to capture contextual information and hidden linguistic pattern without hand-crafted features or regular expressions. The architecture is shown in Figure \ref{fig:model}. %
As illustrated, our DeepVar model contains three parts: 

\noindent\textbf{Input Embeddings.} Each word in the sentence has two types of input: word-level (words in red color) and character-level (characters in green color). For character-level input, we applied one-hot encoding (green circle on top); with respect to word-level input, we used the word embedding (red circle icon; Sec. \ref{sec:related}). It's noted that the word embeddings are pre-trained on a separate large collection of biomedical corpus, while character embeddings are built from our variant BioNER task.

\noindent\textbf{Feature Representation Learning.}  
The character representation (green circle icon) is learned from module with LSTM or CNN ("Char Repr"; Sec. \ref{sec:char_emb}). Then it would be concatenated with word embeddings as the input of word sequence representation learning module (gray boxes in the middle; Sec. \ref{sec:word_learning}). This module contains the stacked BiLSTM networks with residual layer integrated, and it's designed to capture long-term information and effective contextual representations.

\noindent\textbf{Inference Module.} The final word feature representation for each word will be the hidden status from the hidden layer (blue circle). The CRFs inference layer will take it and assign labels to each word (Sec. \ref{sec:inference}). 

\subsection{Character feature Representation}\label{sec:char_emb}

Character information has been proven to be critical for entity identification tasks \cite{chiu2016named,lample2016neural,ma2016end}. 
First of all, character embedding could handle the Out-of-Vocabulary (OOV) words to some extent since it could enclose the morphological similarities to some established words. 
Moreover, it also could be able to insert the orthographic and linguistic patterns for variants such as prefix, suffix, and punctuation. For example, mutation names often contain alphabets, digits, hyphens, and other characters like "HIV-1", "IL2", "rs2297882", and "C$>$T". 
It's crucial to learn all those hidden morphological and orthographic patterns automatically for inference. 

\begin{table}[htbp]
    \caption{The Look-up Table for Character One-hot Encoding} \smallskip
    \centering
    \begin{tabular}{c|c}
        \hline
       letters & abcdefghijklmnopqrstuvwxyz \\ \hline
       digits & 0123456789 \\ \hline
       others & ,;.!?:`'``''/\textbackslash$|$\_@\#\$\%\^{}\&*\~{}+$-$=\textless\textgreater()[]\{\} \\\hline
    \end{tabular}
    \label{tab:lookup}
\end{table}

We represented the character-level input by one-hot encoding through a lookup table. 
The lookup table in our work contains 70 characters, including 26 English letters, ten digits, 33 other characters, and one placeholder for the unknown character. The full list is shown in Table \ref{tab:lookup}. 
Subsequently, each word instance is then represented by a sequence of $m=70$ sized vectors with character sequence length $\mathit{l}$, where $\mathit{l}$ is a hyperparameter in our work. 
Then LSTM or CNN is used to learn character-level representation to capture the hidden morphological and orthographic patterns:

\subsubsection{Character CNN.} \citeauthor{chiu2016named} \shortcite{chiu2016named} and \citeauthor{ma2016end} \shortcite{ma2016end} have investigated the effectiveness of using the CNN structure to encode character sequences. In our work, we employed the same architecture as in \cite{ma2016end}. More specifically, one CNN layer was used following with max-pooling to capture character-level representation. 

\subsubsection{Character BiLSTM.} \citeauthor{lample2016neural} \shortcite{lample2016neural} utilized the BiLSTM to model the global character sequence information. In our work, we employed the same architecture as \cite{lample2016neural} in which final states from the \textit{left-to-right} forward LSTM and \textit{right-to-left} backward LSTM are concatenated as character sequence representations.

\subsection{Word Representation Learning}\label{sec:word_learning}
We employed the BiLSTM in our work to model word-level representations as it's more widely used \cite{lample2016neural,ma2016end,chiu2016named,liu2018empower} and more powerful to capture the contextual distributional sensitivity. 
As shown in the gray boxes of Fig. \ref{fig:model}, our word representation learning module includes n units of modules in which n is a hyper-parameter. Each unit includes 2 BiLSTM layers stacked together, followed by a residual layer. The residual layer would take the hidden states from both BiLSTM layers and apply the transformation.

Basically, the input to an LSTM network is a sequence of vectors $\mathit{X} = \{x_1, x_2, \dots, x_T\}$, where $x_t$ is a representation of a word in the input sentence $x$ at certain layer of the network. The output is a sequence of vectors $\mathit{H} = \{h_1, h_2, \dots, h_T\}$, where $h_t$ is a hidden state vector storing all the useful information at time t. At step t of the recurrent calculation, the network takes $x_t, c_{t-1}, h_{t-1}$ as inputs and produces $c_t, h_t$ through the input ($i_t$), forget ($f_t$) and output ($o_t$) gates via the following intermediate calculations:
\begin{align}\label{eq:lstm}
    \mathbf{i}_t & =  \sigma(\mathbf{W}^i \mathbf{x}_t + \mathbf{U}^i \mathbf{h}_{t-1} + \mathbf{b}^i) 
    \\
    \mathbf{f}_t & = \sigma(\mathbf{W}^f \mathbf{x}_t + \mathbf{U}^f \mathbf{h}_{t-1} + \mathbf{b}^f)
    \\
    \mathbf{o}_t & = \sigma(\mathbf{W}^o \mathbf{x}_t + \mathbf{U}^o \mathbf{h}_{t-1} + \mathbf{b}^o)
    \\
    \mathbf{\hat{c}}_t & = \sigma(\mathbf{W}^c \mathbf{x}_t + \mathbf{U}^g \mathbf{h}_{t_1} + \mathbf{b}^g)
    \\
    \mathbf{c}_t & = \mathbf{f}_t \odot \mathbf{c}_{t-1} + i_t \odot \hat{c}_t
    \\
    \mathbf{h}_t & = \mathbf{o}_t \odot \mathit{tanh}(\mathbf{c}_t)
\end{align}

where $\sigma(\cdot)$ and $\mathit{tanh}(\cdot)$ is the element-wise sigmoid and hyperbolic tangent functions, and $\odot$ denotes element-wise product. $\mathbf{W}^i, \mathbf{W}^f, \mathbf{W}^o, \mathbf{W}^c$ denote the weight matrices of different gates for input $\mathbf{x}_t$, and $\mathbf{U}^i, \mathbf{U}^f, \mathbf{U}^o, \mathbf{U}^c$ are the weight matrices for recurrent hidden state $\mathbf{h}_t$. $\mathbf{b}^i, \mathbf{b}^f, \mathbf{b}^o, \mathbf{b}^c$ denote the bias vectors. As shown in above formulation, we used the LSTM design \cite{hochreiter1997long} without peephole connections. 
A BiLSTM includes forward LSTM and backward LSTM. 
The hidden states of the forward and backward LSTM are concatenated for each word and are the input of the next layer, in our case, the BiLSTM transformation.

The semantic representations learned from a shallow network in \cite{lample2016neural,chiu2016named,ma2016end} isn't able to differentiate the variants apart from genes/proteins having similar orthographic patterns. However, simply increasing the depth of a network won't necessarily improve the performance, and on the contrary, it often leads to a decline in performance beyond a certain point \cite{srivastava2015training}. 
The introduce of residual could bridge some shallow global information to deeper levels, and facilitate to address the vanishing gradient problem when training a deeper network \cite{he2016deep}. In our work, we used the identity residual formulated as: 

\begin{equation}\label{eq:residual}
    y(x) = F(x) + x
\end{equation}
where $F(\cdot)$ is a nonlinear parametric function.

\subsection{Inference Procedure} \label{sec:inference}
The CRFs is commonly used for labeling and segmenting sequences tasks, and also has been extensively applied to NER. It's especially helpful for tasks with strong dependencies between token tags. \citeauthor{reimers2017optimal} \shortcite{reimers2017optimal} and \citeauthor{yang2018design} \shortcite{yang2018design} demonstrated that CRFs can deliver a larger performance increase than the softmax classifier across all NER tasks. \citeauthor{reimers2017optimal} \shortcite{reimers2017optimal} also suggested that a dense layer followed by a linear-chain CRF as variant CRF classifier would be able to maximize the tag probability of the complete sentence. In our work, we employed the same variant CRF classifier design for inference. 

First of all, the output for the sequence from the last residual layer is mapped with a dense layer and a linear chain CRF layer to the number of tags. The linear-chain CRF maximizes the tag probability of the complete sentence. More formally, given an input sentence $x$ of length $N$ $x = [w_1, w_2, \dots, w_N]$ in which $w_t$ is the a word in sentence, we predict corresponding variant types $Y = [y_1, y_2, \dots, y_N]$. The score of a sequence of tags $z$ is defined as:

\begin{equation}\label{eq:probdef}
S(x,y,z) = \sum^{N-1}_{t=1}{\mathcal{T}_{z_{t-1},z_{t}} + \sum^{N}_{t=1}{\mathcal{U}_{x_t,z_{t}}}}
\end{equation}

where $\mathcal{T}$ is a transition matrix in which $\mathcal{T}_{p,q}$ represents the score of transitioning from tag $p$ to tag $q$ and $\mathcal{U}_{x_t,z_{t}}$ represents the score of assigning tag $z$ to word $w$ given representation $x$ at time step $t$. Given the ground truth sequence of tags $z$, we minimize the negative log-likelihood loss function during the training phase:

\begin{align} \label{eq:objective}
    \mathcal{L} & = - log \mathcal{P}(z|x) \nonumber\\
                & = log\sum_{\hat{z}\in\mathcal{Z}}{e^{S(x,y,\hat{z})}} - S(x, y,z)
\end{align}

where $\mathcal{Z}$ is the set of all possible tagging paths. For efficient training and decoding, viterbi algorithm is used.

\section{Experiment Setup}

\subsection{Data}
We trained and tested our model on the expert-annotated corpus from tmVar \cite{wei2013tmvar}, while 20\% of the training is held out for validation. The details are shown in Table \ref{tab:data}.

\begin{table}[htbp]
    \caption{The Data in Our Work} \smallskip
    \centering
    \begin{tabular}{c|r}
        \hline
        Data Set & Size \\ \hline
       Training & 2936 sentences\\
       Validation & 734 sentences\\
       Testing & 1113 sentences \\\hline
    \end{tabular}
    \label{tab:data}
\end{table}

\subsubsection{Tokenization.} \label{sec:processing}
The only preprocessing we performed on the data is customized tokenization.
The traditional tokenization in generic NER tasks would split a sentence by the white space and remove all the digits and special characters. However, those digits and special characters like punctuation are part of the genomic variants. 
Moreover, due to the great variations of those entities, appropriate tokenization 
can significantly affect the performance. For example, whether the sequence ``(lL-2)'' is tokenized to \{``('', ``IL-2'' and ``)''\} or \{``(IL'', ``-'', and ``2)''\} would result in considerable difference in representation learning and accuracy. In our work, we first tokenize a sentence using white space and characters in \{`` '' \# \& \$ \_ * ; $\slash$ \textbackslash $\sim$ ! ? = \{ \} \}, then for each token t, if there's any character from \{``'', . ':\} at the end of t, then strip this character. Finally, strip the brackets if t is bracketed.

\subsubsection{Annotation Scheme.}
The choice of annotation scheme varied from applications. There's no consensus on which one is better. \citeauthor{chiu2016named} \shortcite{chiu2016named} demonstrated that BIOES (for Begin, Inside, Outside, End, Single) could achieve considerable performance improvements over BIO (for Begin, Inside, Outside). \citeauthor{lample2016neural} \shortcite{lample2016neural} showed Using BIOES and BIO yields similar performance. \citeauthor{reimers2017optimal} \shortcite{reimers2017optimal} demonstrated that BIO scheme is preferred over BIOES through extensive experiments on varied NER tasks. Therefore, in our work, we adopted the BIO scheme without comparing it with BIOES or other schemes.

\subsection{Evaluation}

One challenge for NER research is establishing an appropriate evaluation metric \cite{nadeau2007survey}. In particular, entities may be correctly delimited but misclassified, or entity boundaries may be mismatched.
The partial matching (text offsets overlap, e.g., left match or right mach) or oversized boundaries may be considered as accurate tagging in some generic NER tasks. However, same as work in tmVar \cite{wei2013tmvar}, we only considered exact matching (two entities match if their boundaries are identical and tags are correctly classified), and any other prediction was considered as misclassification.

In our work, we use the Precision (P), Recall (R), and macro-averages F1 score (F1) to evaluate different models. 
Moreover, there are three types of variants: DNA mutation, protein mutation, and SNP. Therefore, the tags used for training and prediction are B-DNAMutation, I-DNAMutation, B-ProteinMutation, I-ProteinMuataion, B-SNP (no I-SNP), and O. In the reported results for all DNN models, we removed the tag header B- and I-, and only used the tag body with their entity boundaries to calculate precision, recall and F1 score. 

\begin{table}[htbp]
\caption{Hyperparameters and Training Settings} \smallskip
\label{tab:para}
\centering
\begin{tabular}{c|c}
\hline
Parameters & Values  \\ \hline
max char length & 15, 30, 50\\  
char emb size & 25, 50, 100 \\ 
char emb dropout & 0, 0.25, 0.5 \\ 
char CNN filter size & 30, 50, 70  \\ 
char CNN window & 3, 5, 7  \\ 
char LSTM states & 25, 50, 100  \\ 
char LSTM dropout & 0, 0.25, 0.5 \\ \hline 
max word length & 115 \\ 
word emb & BioW2V, BioELMO \\
        & BioFlair, BioBert \\ 
word repr. learning unit n & 1, 2\\  
word LSTM states & 50, 100, 200\\  
word LSTM dropout & 0, 0.25, 0.5\\ \hline
hidden states & 50, 100, 200\\  
hidden layer dropout & 0, 0.25, 0.5\\  \hline
 batch size & 32, 64, 128\\  
 optimizer & SGD, RMSP, ADAM \\  
 learning rate & 1e-4 \\  
 learning rate decay & 1e-5 \\  
 clipnorm & 1.0 \\  \hline
\end{tabular}
\end{table}

\subsection{Settings}
We implemented our model using Keras with TensorFlow backend. The computations for a single model are run on Tesla P100 GPU. Table \ref{tab:para} summarizes the chosen hyperparameter settings for all DNN models. Moreover, the embedding size for BioW2V is also a hyperparameter, which includes 50 and 100. With respect to the SGD optimizer, besides the common settings, we also set the momentum to 0.9 and used Nesterov. All the models are trained in 100 epochs with early stopping.

\section{Results and Discussion}
In this section, we reported our experimental results and investigated some key components in our model design. We also discussed the relation between genomic variants recognition with other similar NER tasks. 

\subsection{Results}
We compared DeepVar with several state-of-the-art NER systems: (1) generic DNN; (2) vanilla DNN \cite{lample2016neural,ma2016end}; and (3) rule-based machine learning variants identification systems (tmVar and nala). We also investigated both BiLSTM and CNN in learning the character-level representation and compared their role in different models. We performed extensive parameter tuning for the DNN models using settings shown in Table \ref{tab:para}. For vanilla models, we used the same setting as in \cite{lample2016neural,ma2016end} on character feature learning and tuned other settings including the word embedding, word representation, and optimizer. For tmVar \cite{wei2013tmvar} and nala \cite{cejuela2017nala}, we quoted their experimental results directly. It's noted that we used tmVar$^{\mathrm{b}}$ (without post-processing) as baseline while tmVar$^{\mathrm{c}}$ (with post-processing) as state-of-the-art benchmark. The results are reported in Table \ref{tab:rs}. 

First of all, we observed that the DeepVar model achieves significantly higher F1 scores than state-of-the-art vanilla DNN models, nala, and tmVar ($^\mathrm{a,b}$ without post-processing). DeepVar also achieves appreciably higher F1 score than generic DNN models. The DeepVar and generic DNN models differ at the introduction of residual layer, which is designed to learn better semantic representations by training deeper networks. For the results reported in Table \ref{tab:rs}, the generic DNN models achieved best performance using shallow network with single BiLSTM layer. Meanwhile, the result of DeepVar is very close to the best record of tmVar ($^{\mathrm{c}}$ with extensive hand-crafted features and post-processing). However, it's worth noting that DeepVar is a truly end-to-end system requiring no preprocessing, feature engineering, nor post-processing. DeepVar should be able to achieve a higher score by adding the post-processing regex in tmVar and create a more practically useful tool in the real application.

\begin{table}[htbp]
\caption{The Results of Test Set Performance} \smallskip
\label{tab:rs}
\centering
\resizebox{.98\columnwidth}{!}{
\begin{tabular}{c|c|l|l|l}
\hline
Models & Char Repr  & P (\%) & R (\%) & F1 (\%) \\ \hline
\multicolumn{1}{c|}{\begin{tabular}[c]{@{}c@{}}DeepVar \end{tabular}} &
\begin{tabular}[c]{@{}c@{}} BiLSTM \\  CNN\end{tabular} &
\begin{tabular}[c]{@{}c@{}} 91.72 \\  90.67\end{tabular} &
\begin{tabular}[c]{@{}c@{}} 89.86 \\  90.48\end{tabular} &
\begin{tabular}[c]{@{}c@{}} \textbf{90.78} \\  90.58\end{tabular} \\ \hline
\multicolumn{1}{c|}{\begin{tabular}[c]{@{}c@{}}DNN\end{tabular}} & 
\begin{tabular}[c]{@{}c@{}} BiLSTM \\  CNN\end{tabular} &
\begin{tabular}[c]{@{}c@{}} 91.84 \\  90.91\end{tabular} &
\begin{tabular}[c]{@{}c@{}} 89.05 \\  89.25\end{tabular} &
\begin{tabular}[c]{@{}c@{}} 90.42 \\  90.07\end{tabular} \\ \hline
\multicolumn{1}{c|}{\begin{tabular}[c]{@{}c@{}}Vanilla$^{\dag}$\end{tabular}} & 
\begin{tabular}[c]{@{}c@{}} BiLSTM \\ \cite{lample2016neural} \\  CNN \\\cite{ma2016end}\end{tabular} &
\begin{tabular}[c]{@{}c@{}} 88.76 \\ \\ 90.32\end{tabular} &
\begin{tabular}[c]{@{}c@{}} 89.66 \\ \\ 87.02\end{tabular} &
\begin{tabular}[c]{@{}c@{}} 89.20 \\ \\ 88.64\end{tabular} \\ \hline

\multicolumn{2}{l|}{tmVar} & \begin{tabular}[c]{@{}r@{}}85.81\\ 92.01\\ 91.38\end{tabular} & \begin{tabular}[c]{@{}r@{}}80.82\\ 83.72\\ 91.40\end{tabular} & \begin{tabular}[c]{@{}r@{}}83.24$^{\mathrm{a}}$\\ \textbf{87.67$^{\mathrm{b}}$}\\ \textbf{91.39$^{\mathrm{c}}$}\end{tabular} \\ \hline
\multicolumn{2}{l|}{nala}  &
    87.00 & 92.00 & 89.00$^{\mathrm{d}}$\\ \hline

\multicolumn{5}{l}{
    \begin{tabular}[l]{@{}l@{}}
        $^{\dag}$same character learning settings, tuning other settings\\
        $^{\mathrm{a}}$used BIO annotation scheme, no post-processing \\ 
        $^{\mathrm{b}}$used 11 annotation scheme, no post-processing \\ 
        $^{\mathrm{c}}$used 11 annotation scheme, with post-processing \\
        $^{\mathrm{d}}$used partial match, result of exact match should be lower.
    \end{tabular}
}
\end{tabular}
}
\end{table}

\subsection{Word Embeddings}
In our experiments, we compared four different pre-trained domain-specific word embeddings: BioW2V, BioElmo, BioFlair, and BioBert \cite{Beltagy2019SciBERT}. 
More specifically, we used CBOW word2vec \cite{mikolov2013distributed} model to train BioW2V on the large up-to-date collection of PubMed corpus. BioELMO\footnote{\url{https://allennlp.org/elmo}} and BioFlair\footnote{\url{https://github.com/zalandoresearch/flair}} are pre-trained on biomedical literature as well. We used the concatenated representations from the last three layers for BioELMO, and BioFlair took the stacked representations from pubmed-forward and pubmed-backward. BioBert\footnote{\url{https://github.com/allenai/scibert}} was pre-trained on scientific literature, we used the concatenated representations from the last 4 layers.

The best performance of DeepVar is reported on BioW2V, 
however, as shown in Table \ref{tab:embedding}, the overall performances of BioELMO, BioBert, and BioFlair significantly outperform BioW2V in generic DNN models of which were usually built with shallow networks. This interesting observation demonstrated that word2vec can achieve compelling performance in deeper neural networks. Moreover, while the performances of BioBert and BioELMO are very close and slightly better than BioFlair, 
it's surprising that BioBert didn't outperform BioELMO. One suspicious factor is that the BioBert we used was pretrained on scientific articles that contain broader topics than biomedical domains, thus affected by the domain shift problem \cite{komiya2018investigating}\footnote{Recently, \citeauthor{10.1093/bioinformatics/btz682} \shortcite{10.1093/bioinformatics/btz682} published an NCBI BioBert which is pre-trained on PubMed corpus. This BioBert should alleviate the domain shift problem.}.

\begin{table}[htbp]
\caption{The Comparisons on Pre-trained Word Embeddings} \smallskip
\label{tab:embedding}
\centering

\resizebox{.98\columnwidth}{!}{
\begin{tabular}{c|c|l|l|l} 
\hline
Model & Embedding & P (\%) & R (\%) & F1 (\%) \\ \hline
\multicolumn{1}{c|}{\begin{tabular}[c]{@{}c@{}}DeepVar \end{tabular}} &
\begin{tabular}[c]{@{}c@{}} BioW2V \\  BioELMO \\ BioBert \\ BioFlair\end{tabular} &
\begin{tabular}[c]{@{}c@{}} 91.72 \\  90.67 \\ 91.49\\ 91.27\end{tabular} &
\begin{tabular}[c]{@{}c@{}} 89.86 \\  90.48\\ 89.45\\ 89.05\end{tabular} &
\begin{tabular}[c]{@{}c@{}} 90.78 \\  90.58\\ 90.46\\ 90.14\end{tabular} \\ \hline
\multicolumn{1}{c|}{\begin{tabular}[c]{@{}c@{}}DNN\end{tabular}} & 
\begin{tabular}[c]{@{}c@{}} BioW2V \\  BioELMO \\ BioBert \\ BioFlair\end{tabular} &
\begin{tabular}[c]{@{}c@{}} 87.52 \\ 91.84 \\  90.97 \\ 90.22\end{tabular} &
\begin{tabular}[c]{@{}c@{}} 89.47 \\ 89.05 \\  89.86 \\ 89.86\end{tabular} &
\begin{tabular}[c]{@{}c@{}} 88.49 \\ 90.42 \\  90.41 \\ 90.04\end{tabular} \\ \hline
\end{tabular}
}
\end{table}

\subsection{Optimizer}

For DeepVar training, we observed that RMSP slightly outperforms Adam while both of them significantly outperform SGD. For generic DNN models, we had the same observation over RMSP and Adam, while SGD has much worse performance. 
This observation is significantly divergent from knowledge learned from generic NER tasks \cite{lample2016neural,ma2016end,reimers2017optimal,yang2018design} in which SGD and Adam are preferred over RMSP.

\begin{table}[htbp]
\caption{The Comparisons on Optimizers} \smallskip
\label{tab:optm}
\centering
\begin{tabular}{c|c|l|l|l}
\hline
Model & Optimizer & P (\%) & R (\%) &  F1 (\%) \\ \hline
\multicolumn{1}{c|}{\begin{tabular}[c]{@{}c@{}}DeepVar \end{tabular}} &
\begin{tabular}[c]{@{}c@{}} SGD \\  RMSP \\ ADAM \end{tabular} &
\begin{tabular}[c]{@{}c@{}} 87.45 \\ 91.72 \\   91.84\end{tabular} &
\begin{tabular}[c]{@{}c@{}} 85.86\\ 89.86 \\  89.05\end{tabular} &
\begin{tabular}[c]{@{}c@{}} 86.65\\ 90.78 \\  90.42\end{tabular} \\ \hline
\multicolumn{1}{c|}{\begin{tabular}[c]{@{}c@{}}DNN\end{tabular}} & 
\begin{tabular}[c]{@{}c@{}} SGD \\  RMSP \\ ADAM \end{tabular} &
\begin{tabular}[c]{@{}c@{}} 82.52 \\ 91.84 \\  88.36 \end{tabular} &
\begin{tabular}[c]{@{}c@{}} 82.35 \\ 89.05 \\  90.87 \end{tabular} &
\begin{tabular}[c]{@{}c@{}} 82.44 \\ 90.42 \\  89.60 \end{tabular} \\ \hline
\end{tabular}
\end{table}

\subsection{Relation to other NER tasks}
Deep Learning on small datasets is on the horizon of research field. State-of-the-art NER systems and some recent BioNER tasks with large datasets have been discussed in Section \ref{sec:related}. However, it is not uncommon that we may encounter domain-specific applications for which only small datasets are produced with high-quality gold label annotations, due to the cost of expert curation. Besides the genomic variants recognition task in this study, another similar example in the industrial NER domain is personal identifier entity recognition from various products' logs, like user names, passwords, taggers, etc. The machine logging language is different from canonical natural language, and it's hard to read and interpret without domain knowledge. Out of hundreds of thousands of logging records, product experts may only be able to obtain hundreds of samples with gold labels for different types of entities. Those personal identifiers also exhibit quite diverse linguistic heterogeneity as we see in genomic variants, mixed with massive digits and punctuation. Moreover, it is critical to recognize the exact boundaries of personal identifiers and minimize the false negative for preventing privacy leakage. 
Hundreds of rule-based regular expressions are generally developed to capture identifiers, which are painstaking to maintain. It is also hard to generalize them across various products. Our study would motivate such similar NER tasks.

\section{Conclusion}
In this paper, We propose the first end-to-end neural network approach ``DeepVar'' for genomic variant entities identification. The proposed approach significantly outperformed the benchmark baseline and vanilla DNN models. While requiring no feature engineering nor post-processing, it achieved comparable performance to the state-of-the-art rule-based machine learning system. We also demonstrated through detailed analysis that the performance gain is achieved by the introduced residual, which facilitates to train a deeper network, and confirmed the domain-specific contextual word embeddings make significant contributions to the performance gain. The significant reduction of reliance on domain-specific knowledge would play a crucial role in certain expert-costly fields. Our investigation on key components may also shed light upon other deep low-resource NER applications. 

\section{Acknowledgments}
This research was partially supported by Adobe Data Science Research Award.

\bibliography{AAAI-ChengC.9733}
\bibliographystyle{aaai}

\end{document}